\documentclass[a4paper]{llncs}

\usepackage{times}
\usepackage{graphicx}
\usepackage{amsmath}
\usepackage{amssymb}
\usepackage{bm}
\usepackage{booktabs}
\usepackage[pagebackref=false,breaklinks=true,colorlinks,bookmarks=false,linkcolor=blue,citecolor=blue,pdftitle={Comparative Evaluation of Action Recognition Methods via Riemannian Manifolds, Fisher Vectors and GMMs: Ideal and Challenging Conditions},pdfauthor={Johanna Carvajal, Arnold Wiliem, Chris McCool, Brian Lovell, Conrad Sanderson}]{hyperref}

\begin{document}

\title
  {
  Comparative Evaluation of Action Recognition Methods via Riemannian Manifolds, Fisher Vectors and GMMs: Ideal and Challenging Conditions
  }

\author
  {
  Johanna~Carvajal, Arnold~Wiliem, Chris~McCool, Brian~Lovell, Conrad~Sanderson
  }

\institute
  {
  University of Queensland, Brisbane, Australia\\
  Queensland University of Technology, Brisbane, Australia\\
  NICTA, Australia\\
  Data61, CSIRO, Australia\\
  \textcolor{white}{\thanks{Published in: {\bf Lecture Notes in Computer Science (LNCS), Vol. 9794, pp.~88-100, 2016.}}}
  }

\maketitle

\begin{abstract}
\vspace{-6ex}

\begin{sloppypar}
We present a comparative evaluation of various techniques for action recognition while keeping as many variables as possible controlled.
We employ two categories of Riemannian manifolds: symmetric positive definite matrices and linear subspaces. 
For both categories we use their corresponding nearest neighbour classifiers, kernels, and recent kernelised sparse representations.
We compare against traditional action recognition techniques based on Gaussian mixture models and Fisher vectors (FVs).
We evaluate these action recognition techniques under ideal conditions, as well as their sensitivity in more challenging conditions (variations in scale and translation).
Despite recent advancements for handling manifolds, manifold based techniques obtain the lowest performance and their kernel representations are more unstable in the presence of challenging conditions.
The FV approach obtains the highest accuracy under ideal conditions.
Moreover, FV best deals with moderate scale and translation changes.
\end{sloppypar}

\end{abstract}

\section{Introduction}
\vspace{-1ex}

Recently, there has been an increasing interest on action recognition using Riemannian manifolds.  
Such recognition systems can be roughly placed into two main categories: {\bf (i)} based on linear subspaces (LS), and {\bf(ii)} based on symmetric positive definite (SPD) matrices.  
The space of $m$-dimensional LS in $\mathbb{R}^n$ can be viewed as a special case of Riemannian manifolds, known as Grassmann manifolds~\cite{Turaga2011}.

Other techniques have been also applied for the action recognition problem. Among them we can find Gaussian mixture models (GMMs), bag-of-features (BoF), and Fisher vectors (FVs). 
In~\cite{Joha2014b,WeiyaoLin2008} each action is represented by a combination of GMMs and then the decision making is based on the principle of selecting the most probable action according to Bayes' theorem~\cite{Bishop_PRML_2006}.
The FV representation can be thought as an evolution of the BoF representation, encoding additional information~\cite{Perronnin2011}.  
Rather than encoding the frequency of the descriptors for a given video, FV encodes the deviations from a probabilistic version of the visual dictionary (which is typically a GMM)~\cite{Carvajal2016b,HengWang2013}.

Several review papers have compared various techniques for human action recognition~\cite{Aggarwal2011,ShianRu_Ke2013,Poppe2010,Weinland2011,Hassner2013}.
The reviews show how this research area has progressed throughout the years, discuss the current advantages and limitations of the state-of-the-art,
and provide potential directions for addressing the limitations. 
However, none of them focus on how well various action recognition systems work  across same datasets \textit{and} same extracted features. 
An earlier comparison of classifiers for human activity recognition is studied~\cite{Perez2007}. 
The performance comparison with seven classifiers in one single dataset is reported. 
Although this work presents a broad range of classifiers, it fails to provide a more extensive comparison by using more datasets and hence its conclusions may not generalise to other datasets.

So far there has been no systematic comparison of performance between methods based on  SPD matrices and LS.  
Furthermore, there has been no comparison of manifold based methods against traditional action recognition methods based on GMMs and FVs in the presence of realistic and challenging conditions. 
Lastly, existing review papers fail to compare various classifiers using the same features across several datasets. 

\textbf{Contributions.} To address the aforementioned problems, in this work we provide a more detailed analysis of the performance of the aforementioned methods under the same set of features. 
To this end, we test with three popular datasets: KTH~\cite{Schuldt2004}, UCF-Sports~\cite{Rodriguez2008} and UT-Tower~\cite{UT-Tower-Data}. 
For the Riemannian representations we use nearest-neighbour classifiers, kernels  as well as recent kernelised sparse representations.
Finally, we quantitatively show when these methods break and how the performance degrades when the datasets have challenging conditions (translations and scale variations). 
For a fair comparison across all approaches,  we will use the same set of features, as explained in Section~\ref{sub:video_descriptors}.
More specifically, a video is represented as a set of features extracted on a pixel basis. 
We also describe how we use this set of features to obtain both Riemannian features: (1) covariance features that lie in the space of SPD matrices, and (2) LS that lie in the space of Grassmann manifolds.
In Sections~\ref{sec:learning_manifolds} and~\ref{sec:learning_gmm_fv}, we 
summarise learning methods based Riemannian manifolds as well as GMMs and FVs.
The datasets and experiment setup are described in Section~\ref{sec:setup}.
In Section~\ref{sec:experiments}, we present comparative results on three datasets under ideal and challenging conditions.
The main findings are summarised in Section~\ref{sec:conclusions}.

\section{Related Work}
\label{sec:related}
\vspace{-1ex}

Many computer vision applications often lie in non-Euclidean spaces, where the underlying distance metric is not the usual $l_2$ norm~\cite{Vemulapalli2013,Jayasumana2013}. For instance, SPD matrices and LS of the Euclidean space are known to lie on Riemannian manifolds.
Non-singular covariance matrices are naturally SPD~\cite{Alavi2014} and have been used to describe gesture and action recognition in~\cite{Faraki2015_cvpr,Faraki_IET_2015,KaiGuo2013,AndresSanin2013}. 

Grassmann manifolds, which are special cases of  Riemannian manifolds, represent a set of $m$-dimensional linear subspaces and have also been investigated for the action recognition problem~\cite{YuiMan2012,YuiManLui2012,YuiMan2011,OHara2012}. 
The straightforward way to deal with Riemannian manifolds is via the nearest-neighbour (NN) scheme. 
For SPD matrices, NN classification using the log-Euclidean metric for covariance matrices is employed in~\cite{Turaga2010,KaiGuo2013}.
Canonical or principal angles are used as a metric to measure similarity between two LS  and have been employed in conjunction with NN in~\cite{Turaga2010}.

Manifolds can be also mapped to a reproducing kernel Hilbert space (RKHS) by using kernels.
Kernel analysis on SPD matrices and LS has been used for gesture and action recognition in~\cite{Harandi2012,Jayasumana2014,Vemulapalli2013}. 
SPD matrices are embedded into RKHS via a pseudo kernel in~\cite{Harandi2012}. With this pseudo kernel is possible to formulate a locality preserving projections over SPD matrices.
Positive definite radial kernels are used to solve the action recognition problem in~\cite{Jayasumana2014}, where an optimisation algorithm is employed to select the best kernel among the class of positive definite radial kernels on the manifold. 

Recently, the traditional sparse representation (SR) on vectors has been generalised to sparse representations in SPD matrices and LS~\cite{Guo2010_Silhouette,Harandi_eccv_2012,Harandi_iccv_2013,BoyueWang2015}.
While the objective of SR is to find a representation that efficiently approximates elements of a signal class with as few atoms as possible, for the Riemannian SR, any given point can be represented as a sparse combination of dictionary elements~\cite{Harandi_eccv_2012,Harandi_iccv_2013}.
In~\cite{Harandi_iccv_2013}, LS are embedded into the space via isometric mapping, which leads to a closed-form solution for updating a LS representation, atom by atom.
Moreover,~\cite{Harandi_iccv_2013} presents a kernelised version of the dictionary learning algorithm to deal with non-linearity in data.
\cite{Harandi_eccv_2012} outlines the sparse coding and dictionary learning problem for SPD matrices. To this end, SPD matrices are embedded into the RKHS to perform sparse coding.

GMMs have also been explored for the  action detection and classification problems.
Each action is represented by a combination of GMMs in~\cite{WeiyaoLin2008}. 
Each action is modelled by two sets of feature attributes.
The first set represents the change of body size, while the second represents the speed of the action. 
Features with high correlations for describing actions are grouped into the same Category Feature Vector (CFV).
All CFVs related to the same category are then modelled using a GMM.
Other approaches for action recognition using GMM are presented in~\cite{Liangliang2010,Joha2014b}. 
In~\cite{Liangliang2010}, spatio-temporal interest points detectors are used to collect a set of local feature vectors, and each feature vector is modelled via GMMs.
Based on GMMs, the likelihood of each feature vector belonging to a given action of interests can be estimated. 
Actions are modelled  using a GMM using low-dimensional action features in~\cite{Joha2014b}.
For GMMs, the decision making is based on the principle of selecting the most probable action according to Bayes' theorem~\cite{Theodoridis2009}.

Recently, the FV approach has been successfully applied to the action recognition problem~\cite{Carvajal2016b,Oneata2013,HengWang2013}. 
This approach can be thought as an evolution of the BoF representation, encoding additional information~\cite{Perronnin2011,HengWang2013}. 
Rather than encoding the frequency of the descriptors, as for BoF, FV encodes the deviations from a probabilistic version of the visual dictionary.
This is done by computing the gradient of the sample log-likelihood with respect the parameters of the dictionary model. 
Since more information is extracted, a smaller visual dictionary size can be used than for BoF, in order to achieve the same or better performance.

\section{Video Descriptors}
\label{sub:video_descriptors}
\label{sec:features}
\vspace{-1ex}

Here, we describe how to extract from a video a set of features on a pixel level. 
The video descriptor is the same for all the methods examined in this work.
A video ${\mathcal{V}} = \{ {\bm{I}}_t \}_{t=1}^T$ is an ordered set of $T$ frames. Each frame $\bm{I}_t \in \mathbb{R}^ {r\times c}$ can be represented by a set of feature vectors $F_t = \{\bm{f}_p\}_{p=1}^{N_t}$. 
We extract the following $d=14$ dimensional feature vector for each pixel in a given frame~$t$~\cite{Joha2014b}:

\vspace{-1ex}
\begin{equation}\label{eq:features}
\bm{f} = \left[ \; x, \; y, \; \bm{g}, \; \bm{o} \; \right]^\top
\end{equation}

\noindent
where $x$ and $y$ are the pixel coordinates, while $\bm{g}$ and $\bm{o}$ are defined as:

\vspace{-2ex}
\begin{small}
\begin{eqnarray}
\hspace{-2ex} \bm{g} & = & 
\left[ \; |J_x|, \; |J_y|, \; |J_{yy}|, \; |J_{xx}|, \; \sqrt{J_x^2 + J_y^2}, \; \text{atan} \frac{|J_y|}{|J_x|} \; \right]
\label{eq:features2}\\
\hspace{-2ex} \bm{o}  & = & 
\left[\; u, \;\;  v, \;\; \frac{\partial{u}}{\partial{t}}, \;\;  
\frac{\partial{v}}{\partial{t}}, \;\; 
\left (\frac{\partial u}{\partial x} + \frac{\partial v}{\partial y} \right ), \;\;
\left (\frac{\partial v}{\partial x} - \frac{\partial u}{\partial y} \right ) \;
\right]
\label{eq:features3}
\end{eqnarray}%
\end{small}%

The first four gradient-based features in (\ref{eq:features2}) represent the first and second order intensity gradients at pixel location $(x,y)$.
The last two gradient features represent gradient magnitude and gradient orientation.
The optical flow based features in (\ref{eq:features3}) represent:
the horizontal and vertical components of the flow vector,
the first order derivatives with respect to time,
the divergence and vorticity of optical flow~\cite{Ali2010}, respectively.
Typically only a subset of the pixels in a frame correspond to the object of interest ($N_t < r \times c$).
As such, we are only interested in pixels with a gradient magnitude greater than a threshold~$\tau$~\cite{KaiGuo2013}.
We discard feature vectors from locations with a small magnitude,
resulting in a variable number of feature vectors per frame.

For each video ${\mathcal{V}}$, the feature vectors are pooled into set $\mathcal{F}=\{\bm{f}_n \}_{n=1}^N$ containing $N$ vectors. 
This pooled set of features $\mathcal{F}$ can be used directly by methods such as GMMs and FVs.
Describing these features using a Riemannian Manifold setting requires a further step to produce either a covariance matrix feature or a linear subspace feature.

Covariance matrices of features have proved very effective for action recognition \cite{KaiGuo2013,AndresSanin2013}. 
The empirical estimate of the covariance matrix of set $\mathcal{F}$ is given by:

\vspace{-1ex}
\begin{equation}\label{eq:cov_features}
\bm{C}= \frac{1}{N} \sum\nolimits_{n=1}^N \left(\bm{f}_n- \overline{\mathcal{F}} \right)\left(\bm{f}_n- \overline{\mathcal{F}} \right)^\top
\end{equation}

\noindent where $\overline{\mathcal{F}}=\frac{1}{N}\sum_{n=1}^N \bm{f}_n$ is the mean feature vector. 

The pooled feature vectors set $\mathcal{F}$  can be represented as a linear subspace through any orthogonalisation procedure like singular value decomposition (SVD)~\cite{Mehrtash2013}. Let $\mathcal{F}=\bm{UDV}^\top$ be the SVD of $\mathcal{F}$. The first $m$ columns of $\bm{U}$ represent an optimised subspace of order $m$. 
The Grassmann manifold ${\cal{G}}_{d,m}$ is the set of $m$-dimensional linear subspaces of $\mathbb{R}^d$. An element of ${\cal{G}}_{d,m}$ can be represented by an orthonormal matrix $\bm{Y}$ of size $d \times m$ such that $\bm{Y}^\top\bm{Y} =\bm{I}_m$, where $\bm{I}_m$ is the $m \times  m$ identity matrix.

\section{Classification on Riemannian Manifolds}
\label{sec:learning_manifolds}

\subsection{Nearest-Neighbour Classifier}

The Nearest Neighbour (NN) approach classifies a query data based on the most similar observation in the annotated training set~\cite{Narasimha2011}. To decide whether two observations are similar we will employ two metrics: the log-Euclidean distance for SPD matrices~\cite{KaiGuo2013} and the Projection Metric for LS~\cite{Hamm2008}. 

The log-Euclidean distance ($d_\text{spd}$) is one of the most popular metrics for SPD matrices due to its accuracy and low computational complexity~\cite{Arsigny2006}; it is defined as:

\vspace{-1ex}
\begin{equation}
\label{eq:logEucl_dist}
d_\text{spd}(\bm{C}_1,\bm{C}_2) = || \log(\bm{C}_1) -\log(\bm{C}_2)||_F
\end{equation}

\noindent where $\log(\cdot)$ is the matrix-logarithm and $||\cdot||_F$ denotes the Frobenius norm on matrices. 

As for LS, a common metric to measure the similarity between two subspaces is via principal angles~\cite{Hamm2008}. The metric can include the smallest principal angle, the largest principal angle, or a combination of all principal angles~\cite{Hamm2008,Vemulapalli2013}. In this work we have selected the Projection Metric which uses all the principal angles~\cite{Hamm2008}:

\vspace{-1ex}
\begin{equation}
d_\text{ls} (\bm{Y}_1,\bm{Y}_2) =  \left ( m - \sum\nolimits_{i=1}^m \cos^2\theta_i \right) ^{1/2}
\label{eq:PM}
\end{equation}

\vspace{-1ex}
\noindent where $m$ is the size of the subspace. 

The principal angles can be easily computed from the SVD of {\small $\bm{Y}_1^\top\bm{Y}_2 \mbox{~=~} \bm{U}(\cos \Theta)\bm{V}^\top$}, where $\bm{U} = [\bm{u}_1 \cdots \bm{u}_m]$, $\bm{V} = [\bm{v}_1 \cdots \bm{v}_m]$, and $\cos \Theta = \textrm{diag} (\cos \theta_1, \cdots, \cos \theta_m)$. 

\subsection{Kernel Approach}

Manifolds can be mapped to Euclidean spaces using Mercer kernels~\cite{Vemulapalli2013,Harandi_eccv_2012,Jayasumana2013}. 
This transformation allows us to employ algorithms originally formulated for $\mathbb{R}^n$ with manifold value data. 
Several  kernels for the set of SPD matrices have been proposed in the literature \cite{Harandi_eccv_2012,Jayasumana2013,JianjiaZhang2104}.
One kernel based on the log-Euclidean distance is derived in~\cite{R_Wang2012} and various kernels can be generated, including~\cite{Vemulapalli2013}:

\vspace{-1ex}
\noindent
\begin{small}
\begin{eqnarray}
K_\text{spd}^\text{rbf} (\bm{C}_1,\bm{C}_2) & = & \exp\left(-\gamma_r \cdot || \log(\bm{C}_1) - \log(\bm{C}_2)||^2_F\right) \label{eq:LED_RBF_ker}\\
K_\text{spd}^\text{poly} (\bm{C}_1,\bm{C}_2) & = & \left(\gamma_p \cdot \text{tr}\left[\log(\bm{C}_1)^\top \log(\bm{C}_2)\right]\right)^d  \label{eq:LED_Poly_ker}
\end{eqnarray}
\end{small}

Similar to SPD kernels, many kernels have been proposed for LS~\cite{Mehrtash2013,Sareh2012}. Various kernels can be generated from the projection metric, such as~\cite{Vemulapalli2013}:

\vspace{-1ex}
\noindent
\begin{small}
\begin{eqnarray}
K_\text{ls}^\text{rbf} (\bm{Y}_1,\bm{Y}_2) & = & \exp\left(-\gamma_r \cdot || \bm{Y}_1\bm{Y}_1^\top - \bm{Y}_2\bm{Y}^\top_2 ||^2_F \right) \label{eq:ProjRBF_ker} \\
K_\text{ls}^\text{poly} (\bm{Y}_1,\bm{Y}_2) & = & \left(\gamma_p \cdot || \bm{Y}^\top_1\bm{Y}_2 ||^2_F\right)^m  \label{eq:ProjPoly_ker}
\end{eqnarray}
\end{small}

\noindent The parameters $\gamma_r$ and $\gamma_p$ are defined in Section~\ref{sec:setup}.
The kernels are used in combination with Support Vector Machines (SVMs)~\cite{Bishop_PRML_2006}.

\subsection{Kernelised Sparse Representation}

Recently, several works show the efficacy of sparse representation methods for addressing manifold feature classification problems~\cite{YuweiWu_img_pro_2015,Harandi_iccv_2013}. Here, each manifold point is represented by its sparse coefficients.
Let $\mathcal{X} = \{ \bm{X}_j\}_{j=1}^J$ be a population of Riemannian points (where $\bm{X}_j$ is either a SPD matrix or a LS)  and ${\mathcal{D}=\{\bm{D}_i}\}_{i=1}^K$ be the Riemannian dictionary of size $K$,  where each element represents an atom. Given a kernel $k(\cdot,\cdot)$, induced by the feature mapping function $\phi: \mathbb{R}^{d}\rightarrow \mathbb{H}$, we seek to learn a dictionary and corresponding sparse code $\bm{s} \in \mathbb{R}^{K}$ such that  $\phi(\bm{X})$ can be well approximated by the dictionary $\phi(\mathcal{D})$. The kernelised dictionary learning in Riemannian manifolds optimises the following objective function~\cite{YuweiWu_img_pro_2015,Harandi_iccv_2013}:

\vspace{-2ex}
\begin{equation}
\displaystyle\min_{\bm{s}} \left( \left|\left| \phi (\bm{X} )
 - \sum\nolimits_{i=1}^K s_i  \phi (\bm{D}_i) \right|\right|_F^{2} 
 + \lambda \left|\left| \bm{s} \right|\right|_1\right)
\end{equation}
\vspace{-1ex}

\noindent over the dictionary and the sparse codes $\mathcal{S} = \{ \bm{s}_j\}_{j=1}^J$.
After initialising the dictionary $\mathcal{D}$, the objective function is solved by repeating two steps (sparse coding and dictionary update).
In the sparse coding step, $\mathcal{D}$ is fixed and $\mathcal{S}$ is computed.
In the dictionary update step, $\mathcal{S}$ is fixed while $\mathcal{D}$ is updated, with each dictionary atom updated independently.

For the sparse representation on SPD matrices, each atom $\bm{D}_i \in \mathbb{R}^{d\times d}$ and each element $\bm{X} \in \mathbb{R}^{d\times d}$ are SPD matrices. The dictionary is learned following~\cite{Harandi_eccv_2012}, where the dictionary is initialised using the Karcher mean~\cite{Bini2013}.
For the sparse representation on LS, the dictionary   $\bm{D}_i \in \mathbb{R}^{d\times m}$ and  each element $\bm{X} \in \mathbb{R}^{d\times m}$ are elements of $\mathcal{G}_{d,m}$ and need to be determined by the Kernelised Grassmann Dictionary Learning  algorithm proposed in~\cite{Harandi_iccv_2013}.
We refer to the kernelised sparse representation (KSR) for SPD matrices and LS as KSR$_{\text{spd}}$ and KSR$_{\text{ls}}$, respectively.

\section{Classification via Gaussian Mixture Models and Fisher Vectors}
\label{sec:learning_gmm_fv}
\vspace{-1ex}

A {\bf Gaussian Mixture Model} (GMM) is a weighted sum of $K$ component Gaussian densities~\cite{Bishop_PRML_2006}:

\vspace{-3ex}
\begin{equation}
p(\bm{f} | \lambda) = \sum\nolimits_{k=1}^{K}w_k\mathcal{N}(\bm{f} | \bm{\mu}_k, \bm{\Sigma}_k)
\end{equation}

\noindent
where $\bm{f}$ is a $d$-dimensional feature vector, $w_k$ is the weight of the $k$-th Gaussian (with constraints $0\leq w_k \leq 1$ and $\sum_{k=1}^{K}w_k=1$), and $\mathcal{N}(\bm{f} | \bm{\mu}_k, \bm{\Sigma}_k)$ is the component Gaussian density with mean $\bm{\mu}$ and covariance matrix $\bm{\Sigma}$, given by: 

\vspace{-1ex}
\begin{small}
\begin{equation}
\mathcal{N}(\bm{f} | \bm{\mu}, \bm{\Sigma}) = \frac{1}{(2\pi)^{\frac{d}{2}}|\bm{\Sigma}|^{\frac{1}{2}}}
\exp {\left\{ -\frac{1}{2}(\bm{f}-\bm{\mu})^\top\bm{\Sigma}^{-1}(\bm{f}-\bm{\mu}) \right\}}
\end{equation}%
\end{small}%

The complete Gaussian mixture model is parameterised by the mean vectors, covariance matrices and weights
of all component densities. These parameters are collectively represented by the notation
$\lambda= \{w_k,\bm{\mu}_k,\bm{\Sigma}_k\}_{k=1}^{K}$.
For the GMM, we learn one model per action. This results in a set of GMM models that we will express as $\{ \lambda_a \}_{a=1}^A $, where $A$ is the total number of actions. 
For each testing video ${\mathcal{V}}$, the feature vectors in set $\mathcal{F}$  are assumed independent, so the average log-likelihood of a model $\lambda_a$ is computed as:

\vspace{-1ex}
\begin{equation}\label{eq:log_like}
\log p(\mathcal{F}|\lambda_a) = \frac{1}{N}\sum\nolimits_{n=1}^N  \log p(\bm{f}_n|\lambda_a)
\end{equation}

\noindent
We classify each video to the model $a$ which has the highest average log-likelihood.

The {\bf Fisher Vector} (FV) approach encodes the deviations from a probabilistic visual dictionary,
which is typically a GMM with diagonal covariance matrices~\cite{JorgeSanchez2013}.
The parameters of a GMM with $K$ components can be expressed as $\lambda=\{w_k,\bm{\mu}_k,\bm{\sigma}_k\}_{k=1}^{K}$,
where, $w_k$ is the weight, $\bm{\mu}_k$ is the mean vector, and $\bm{\sigma}_k$ is the diagonal covariance matrix
for the $k$-th Gaussian.
The parameters are learned using the Expectation Maximisation algorithm~\cite{Bishop_PRML_2006} on training data.
Given the pooled set of features $\mathcal{F}$ from video ${\mathcal{V}}$, the deviations from the GMM are then accumulated using~\cite{JorgeSanchez2013}:

\vspace{-2ex}
\begin{small}
\begin{eqnarray}
\mathcal{G}_{\bm{\mu}_{k}}^{\mathcal{F}}    & = & \frac{1}{N\sqrt{w_k}} \sum\nolimits_{n=1}^{N} \gamma_n(k)\left( \frac{\bm{f}_n - \bm{\mu}_k}{\bm{\sigma}_k} \right)\\
\mathcal{G}_{\bm{\sigma}_{k}}^{\mathcal{F}} & = & \frac{1}{N\sqrt{2w_k}} \sum\nolimits_{n=1}^{N} \gamma_n(k)\left[ \frac{\left(\bm{f}_n - \bm{\mu_k}\right)^2}{\bm{\sigma}_k^2} -1 \right]
\end{eqnarray}%
\end{small}%

\noindent where vector division indicates element-wise division
and $\gamma_n(k)$ is the posterior probability of $\bm{f}_n$ for the $k$-th component:

\vspace{-2ex}
\begin{equation}
\gamma_n(k) = \frac{w_k\mathcal{N}(\bm{f}_n|\bm{\mu}_k, \bm{\sigma}_k)}{\sum\nolimits_{i=1}^{K}w_i\mathcal{N}(\bm{f}_n|\bm{\mu}_i, \bm{\sigma}_i)}
\end{equation}%

\noindent
The Fisher vector for each video ${\mathcal{V}}$ is represented as the concatenation of
{\small $\mathcal{G}_{\bm{\mu}_{k}}^{\mathcal{F}}$} and {\small $\mathcal{G}_{\bm{\sigma}_{k}}^{\mathcal{F}}$} (for {\small $k$~=~$1, \ldots, K$}) into vector~$\mathcal{G}_{\lambda}^{\mathcal{F}}$.
As {\small $\mathcal{G}_{\bm{\mu}_{k}}^{\mathcal{F}}$} and {\small $\mathcal{G}_{\bm{\sigma}_{k}}^{\mathcal{F}}$} are \mbox{$d$-dimensional},
$\mathcal{G}_{\lambda}^{\mathcal{F}}$ has the dimensionality of $2dK$.
Power normalisation is then applied to each dimension in $\mathcal{G}_{\lambda}^{\mathcal{F}}$.
The power normalisation to improve the FV for classification was proposed in~\cite{Perronnin2010} of the form $z \leftarrow \text{sign}(z)|z|^\rho$,  where $z$ corresponds to each dimension and the power coefficient $\rho =1/2$. Finally, $l_2$-normalisation is applied.
Note that we have omitted the deviations for the weights as they add little information~\cite{JorgeSanchez2013}.
The FVs are fed to a linear SVM for classification, where the similarity between vectors is measured using dot-products~\cite{JorgeSanchez2013}.

\section{Datasets and Setup}
\label{sec:setup}

\begin{figure}[!tb]
  \begin{minipage}{1\textwidth}
    \centering
    \begin{minipage}{0.15\textwidth}
      KTH
    \end{minipage}
    \begin{minipage}{0.6\textwidth}
      \centering
      \includegraphics[height=0.22\textwidth,width=0.32\textwidth]{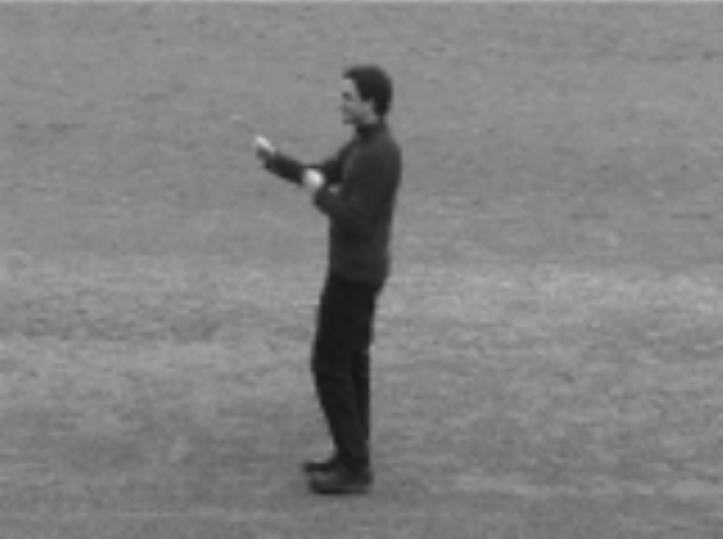}
      \includegraphics[height=0.22\textwidth,width=0.32\textwidth]{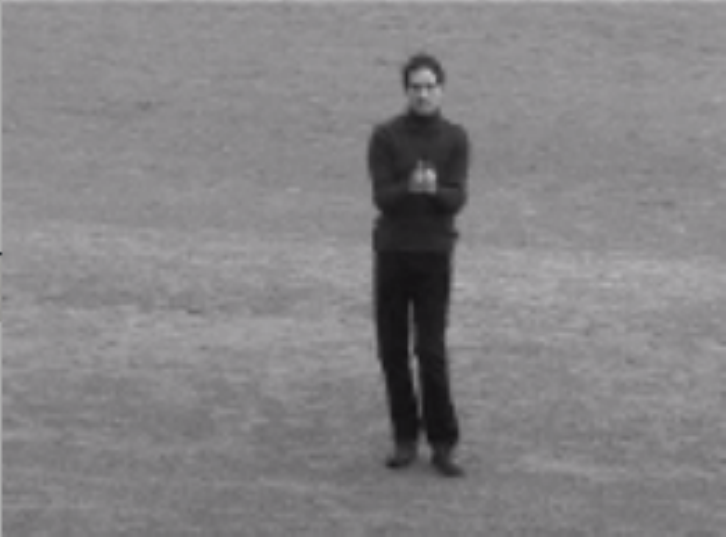}
      \includegraphics[height=0.22\textwidth,width=0.32\textwidth]{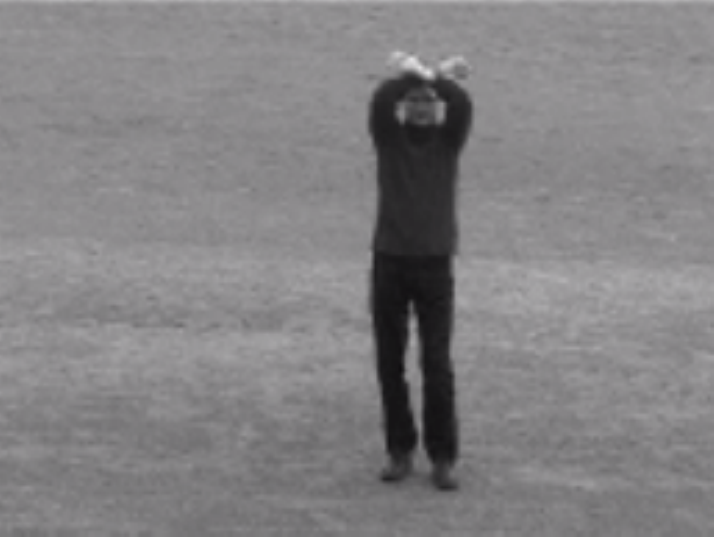}
    \end{minipage}
  \end{minipage}
  \begin{minipage}{1\textwidth}
    \centering
    \begin{minipage}{0.15\textwidth}
      UCF-Sports
    \end{minipage}
    \begin{minipage}{0.6\textwidth}
      \centering
      \includegraphics[height=0.22\textwidth,width=0.32\textwidth]{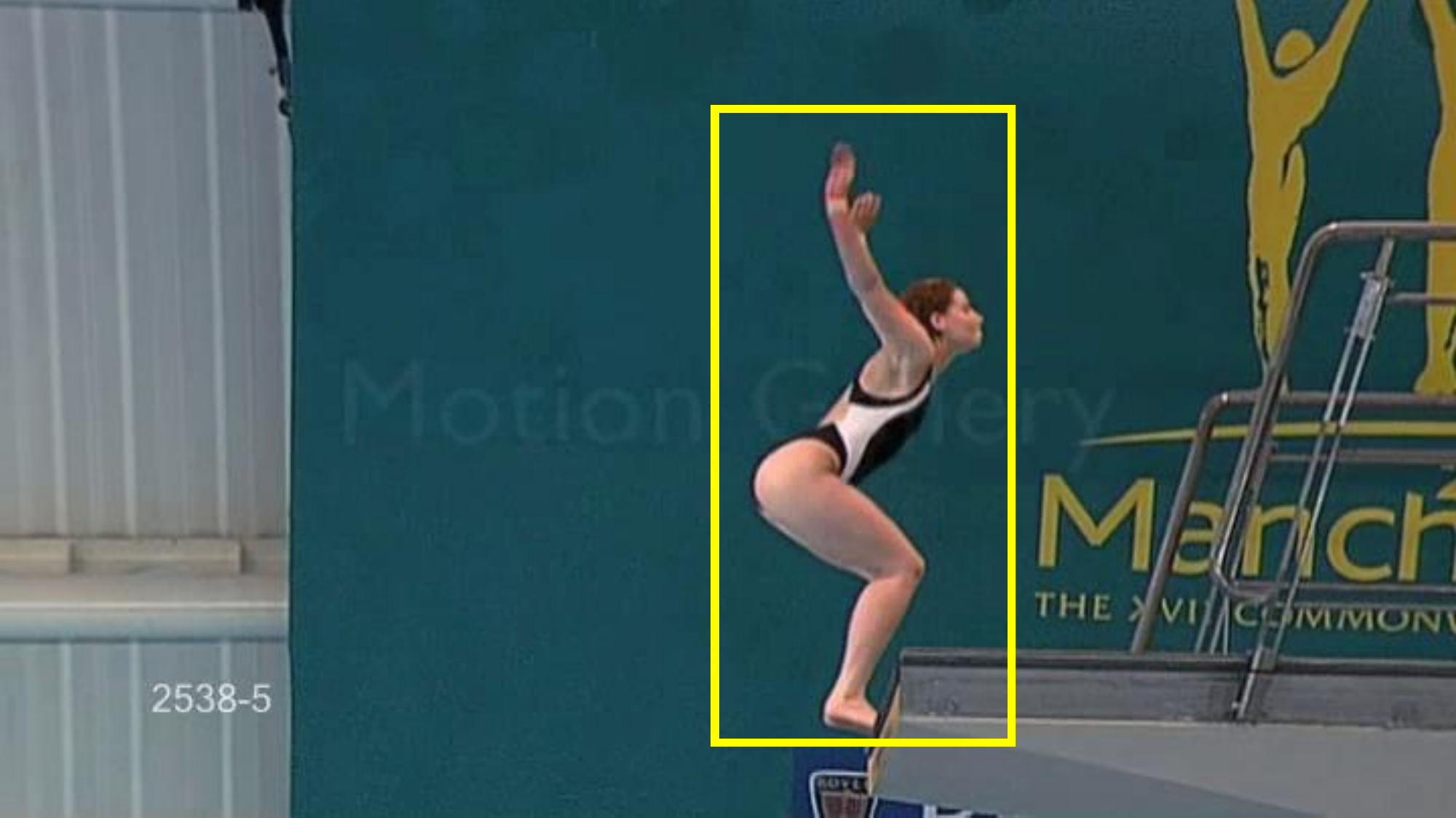}
      \includegraphics[height=0.22\textwidth,width=0.32\textwidth]{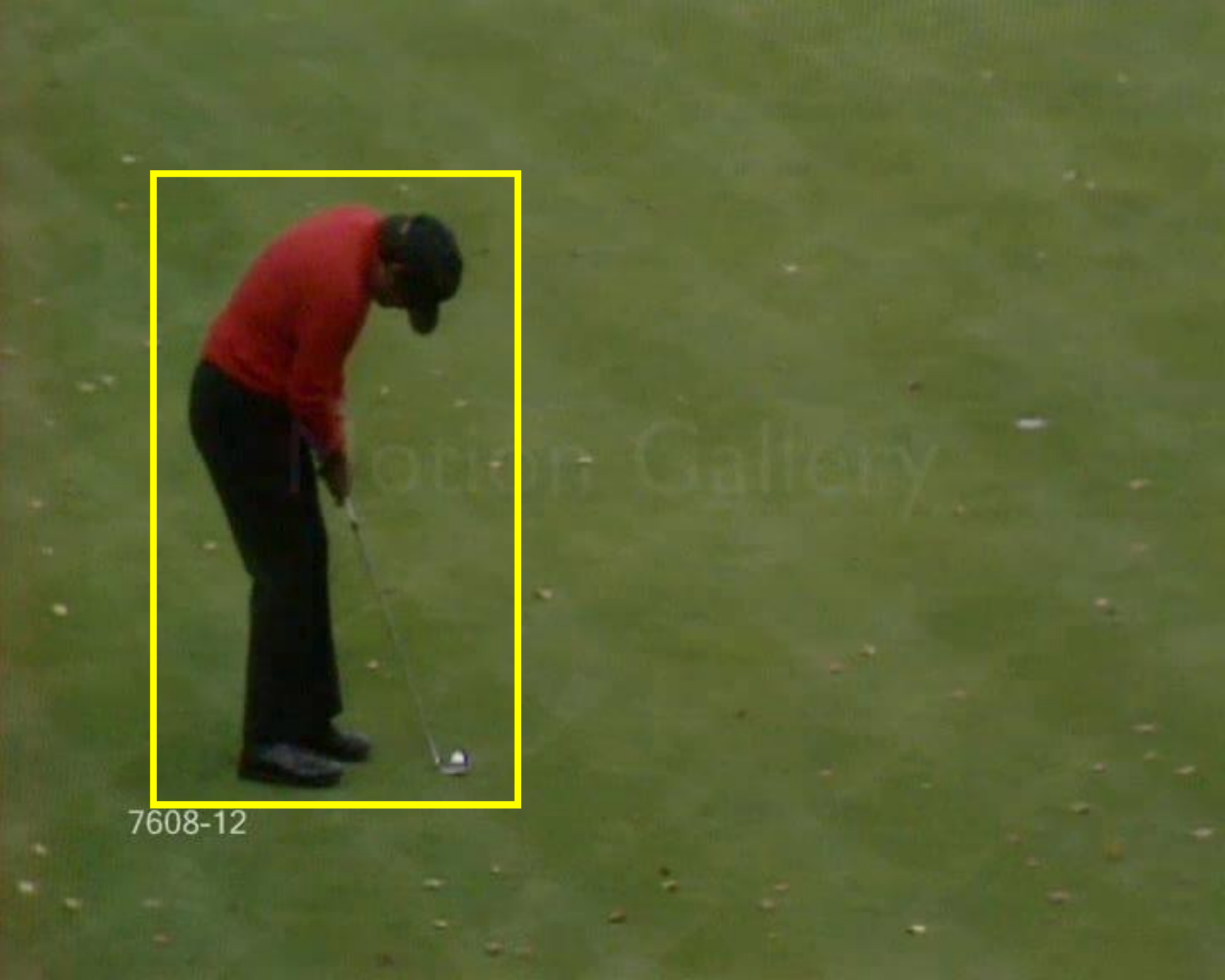}
      \includegraphics[height=0.22\textwidth,width=0.32\textwidth]{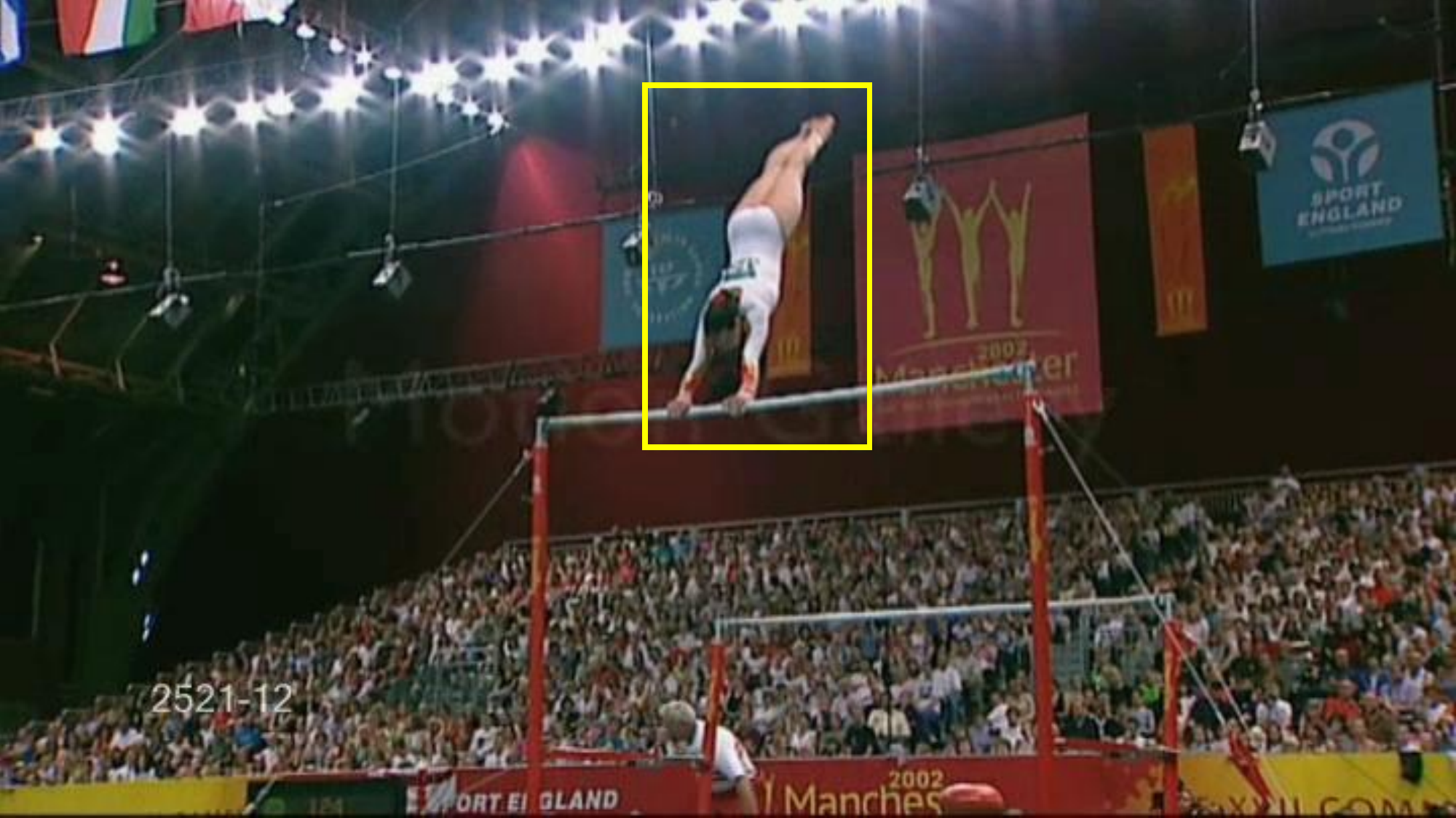}\\
    \end{minipage}
  \end{minipage}
  \begin{minipage}{1\textwidth}
    \centering
    \begin{minipage}{0.15\textwidth}
      UT-Tower
    \end{minipage}
    \begin{minipage}{0.6\textwidth}
      \centering
      \includegraphics[height=0.22\textwidth,width=0.32\textwidth]{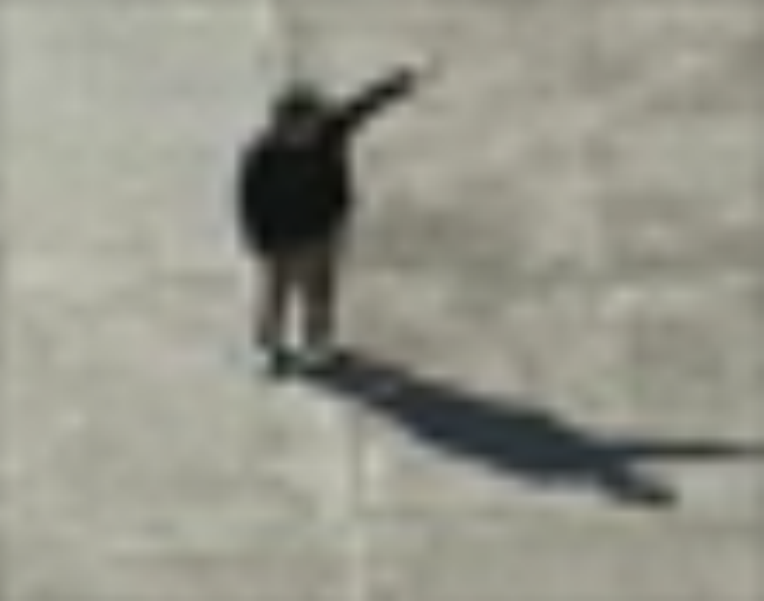}
      \includegraphics[height=0.22\textwidth,width=0.32\textwidth]{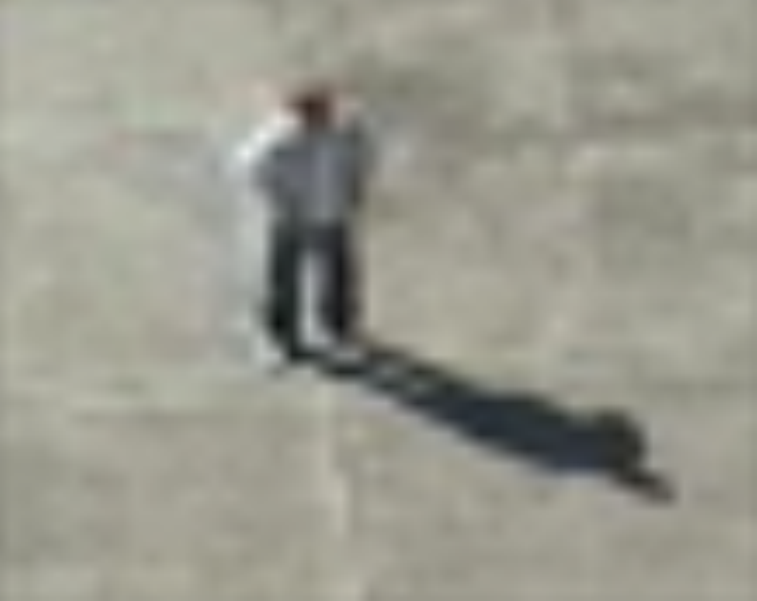}
      \includegraphics[height=0.22\textwidth,width=0.32\textwidth]{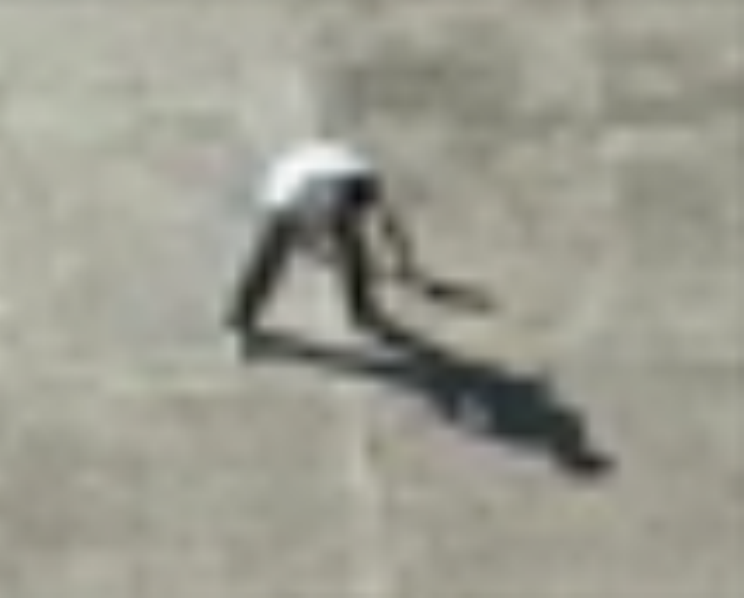}
    \end{minipage}
  \end{minipage}
  \caption{Examples from the three datasets.}
  \label{fig:datasets_examples}
  \vspace{-2ex}
\end{figure}

For our experiments, we use three datasets: KTH~\cite{Schuldt2004}, UCF-Sports~\cite{Rodriguez2008}, and UT-Tower~\cite{UT-Tower-Data}. 
See Fig.~\ref{fig:datasets_examples} for examples of actions. 
In the following sections, we describe how each of the datasets is employed, and we provide a description of the setup used for the experiments.

{\bf Datasets.}
The {KTH} dataset~\cite{Schuldt2004}  contains 25 subjects performing 6 types of human actions and 4 scenarios. The actions included in this dataset are: boxing, handclapping, handwaving, jogging, running, and walking.
The scenarios include indoor, outdoor, scale variations, and varying clothes.
Each original video of the KTH dataset contains an individual performing the same action.  
The image size is 160$\times$120 pixels, and temporal resolution is 25 frames per second. For our experiments we only use scenario 1. 

The {UCF-Sports} dataset~\cite{Rodriguez2008} is a collections of 150 sport videos or sequences. 
This datasets consists of $10$ actions: diving, golf swinging, kicking a ball, lifting weights, riding horse, running, skate boarding, pommel horse, high bar, and walking.  
The number of videos per action varies from $6$ to $22$.  
The videos presented in this dataset have varying backgrounds. 
We use the bounding box enclosing the person of interest provided with the dataset where available. 
We create the corresponding 10 bounding boxes not provided with the dataset. The bounding box size is 250$\times$400 pixels.  

The {UT-Tower} dataset~\cite{UT-Tower-Data} contains $9$ actions performed $12$ times. 
In total, there are 108 low-resolution videos. The actions include: pointing, standing, digging, walking, carrying, running, wave1, wave2, and jumping. The videos were recorded in two scenes (concrete square and lawn). 
As the provided bounding boxes have variable sizes, we resize the resulting  boxes to \mbox{\small $32\times 32$}.

{\bf Setup.}
We use the Leave-One-Out (LOO) protocol suggested by each dataset. We leave one sample video out for testing on a rotating basis for UT-Tower and UCF-Sports. For KTH we leave one person out.
For each video we extract a set of $d=14$ dimensional features vectors as explained in Section~\ref{sec:features}. 
We only use feature vectors with a gradient magnitude greater that a threshold $\tau$. 
The threshold $\tau$ used for selecting low-level feature vectors was set to 40 as per~\cite{Joha2014b}.

For each video, we obtain one SPD matrix and one LS.  
In order to obtain the optimised linear subspace ${\cal{G}}_{d,m}$ in the the manifold representation, we vary $m=1,\cdots, d$. 
We test with manifolds kernels using various parameters. 
The set of parameters was used as proposed in ~\cite{Vemulapalli2013}.
Polynomial kernels $K_\text{spd}^\text{poly}$  and   $K_\text{ls}^\text{rbf}$ are generated by taking $\gamma_p = 1/d_p$ and $d_p=\{1,2,\cdots,d\}$. 
Projection RBF kernels are generated with $\gamma_r=\frac{1}{d}2^{\delta}$ and $\delta=\{-10,-9,\cdots,9\}$ for $K_\text{spd}^\text{rbf}$, and $\delta=\{-14,-12,\cdots,20\}$ for $K_\text{ls}^\text{rbf}$.
For the sparse representation of SPD matrices and LS we have used the code provided by~\cite{Harandi_eccv_2012,Harandi_iccv_2013}. 
Kernels are used in combination with SVM for final classification.
We report the best accuracy performance after iterating with various parameters. 

For the FV representation, we use the same set-up as in~\cite{HengWang2013}. We randomly sampled 256,000 features from training videos and then the visual dictionary is learned with $256$ Gaussians. 
Each video is represented by a FV.
The FVs are fed to a linear SVM for classification.
For the GMM  modelling, we learn a model for each action using all the feature vectors belonging to the same action. For each action a GMM is trained with $K=256$ components. 
The experiments were implemented with the aid of the Armadillo C++ library~\cite{Armadillo2016}.

\vspace{-1ex}
\section{Comparative Evaluation}
\label{sec:experiments}
\vspace{-2ex}

We perform two sets of experiments: {\bf (i)} in ideal conditions, where the classification is carried out using each original dataset, and {\bf (ii)} in realistic and challenging conditions where testing videos are modified by scale changes and translations.

\vspace{-1ex}
\subsection{Ideal Conditions}
\label{sec:ideal_cond}
\vspace{-1ex}

We start our experiments using the NN classifier for both Riemannian representations: SPD matrices and LS. 
For LS we employ the projection metric as per Eq.~(\ref{eq:PM}) and for SPD matrices we employ the log-Euclidean distance as per Eq.~(\ref{eq:logEucl_dist}). We tune the parameter $m$ (subspace order) for each dataset. 
The kernels selected for SPD matrices and LS are described in Eqs.~(\ref{eq:LED_RBF_ker})-(\ref{eq:ProjPoly_ker}) and their parameters are selected as per Section~\ref{sec:setup}. 

We present a summary of the best performance obtained for the manifold representations using the optimal subspace for LS and also the optimal kernel parameters for both representations. 
Similarly, we report the best accuracy performance for the kernelised sparse representations {KSR$_{\text{spd}}$}	 and  {KSR$_{\text{ls}}$}.
Moreover, we include the performance for the GMM and FV representations. 

\begin{table}[!tb]
\centering
\caption {Accuracy of action recognition in ideal conditions.} \label{tab:results} 
\vspace{-1ex}
\begin{tabular}{ccccc} \toprule
     						 					& {~~~~~~KTH~~~~~~} 		& {~UCF-Sports~}  & {~UT-Tower~} & {~~~average~~~}  \\ \midrule
$d_\text{spd}$~\scriptsize{+ NN}					&  $76.0\%$	& $76.5\%$ & $73.1\%$ & $75.2\%$ \\
$d_\text{ls}$~~\scriptsize{+ NN}					&  $67.3\%$ 	& $65.7\%$ & $76.8\%$ & $69.9\%$  \\
\midrule
$K_\text{spd}^\text{poly}$~\scriptsize{+ SVM}	&  $92.0\%$ 	& $75.2\%$ & $87.9\%$ & \underline{$85.0\%$} 
    \vspace{2mm}\\
$K_\text{spd}^\text{rbf}$~\scriptsize{+ SVM}		&  $84.0\%$ 	& $79.2\%$ & $81.5\%$ & $81.6\%$ 
\vspace{2mm}\\
$K_\text{ls}^\text{poly}$~\scriptsize{+ SVM}		& $56.0\%$	& $50.3\%$ & $42.6\%$& $49.6\%$ 
\vspace{2mm}\\
$K_\text{ls}^\text{rbf}$~\scriptsize{+ SVM}		& $76.0\%$ 	& $61.7\%$ &$79.6\%$ & $72.4\%$ \\
\midrule
{KSR$_{\text{spd}}$}~\scriptsize{+ SVM}			&  $80.0\%$ 	& $76.5\%$ & $81.5\%$
\vspace{1mm} & $79.3\%$ \\
{KSR$_{\text{ls}}$}~\scriptsize{+ SVM}			&  $74.0\%$ 	& $72.5\%$ & $83.3\%$ & $77.3\%$ \\
\midrule
{GMM}					 						&  $86.7\%$ 	& $80.5\%$ & $87.9\%$ & \underline{$85.0\%$} \\
\midrule
{FV}	~\scriptsize{+ SVM}							&  $\bm{96.7\%}$ & $\bm{88.6\%}$ & $\bm{92.5\%}$ & \underline{$\bm{92.6}\%$} \\ \bottomrule
\end{tabular}
\end{table}

The results are presented in Table~\ref{tab:results}.
First of all, we observe that using a SVM for action recognition usually leads to a better accuracy than NN.
In particular, we notice that the NN approach performs quite poorly.
The NN classifier may not be effective enough to capture the complexity of the human actions when there is insufficient representation of the actions (one video is represented by one SPD matrix or one LS).
Secondly, we observe that among the manifold techniques, SPD based approaches perform better than LS based approaches.
While LS capture only the dominant eigenvectors~\cite{Sareh2015}, SPD matrices capture both the eigenvectors and eigenvalues~\cite{Traore2011}. The eigenvalues of a covariance matrix typify the variance captured in the direction of each eigenvector~\cite{Traore2011}. 

Despite KSR$_{\text{spd}}$ showing superior performance in other computer vision tasks~\cite{Harandi_eccv_2012}, it is not the case for the action recognition problem. 
We conjecture this is due to the lack of labelled training data (each video is represented by only one SPD matrix), which may yield a dictionary with bad generalisation power. 
Moreover, sparse representations can be over-pruned, being caused by discarding  several representative points that may be potentially useful for prediction~\cite{Hirose2015}.

Although kernel approaches map the data into higher spaces to allow linear separability, $K_\text{spd}^{poly}$ exhibits on average a similar accuracy to GMM  which does not transform the data. 
GMM is a weighted sum of Gaussian probability densities, which in addition to the covariance matrices, it uses the means and weights to determine the average log-likelihood of a set of feature vectors from a video to  belong to a specific action. 
While SPD kernels only use covariance matrices, GMMs use both covariance matrices and means. 
The combination of both statistics has proved to increase the accuracy performance in other classification tasks~\cite{Aggarwal2012,JorgeSanchez2013}.
FV outperforms all the classification methods with an average accuracy of  $92.6\%$, which is $7.6$ points higher than both GMM and $K_\text{spd}^{poly}$.
Similarly to GMM, FV also incorporates first and second order statistics (means and covariances), but it has additional processing in the form of power normalisation. 
It is shown in~\cite{Perronnin2010} that when the number of Gaussians increases, the FV turns into a sparser representation and it negatively affects the linear SVM which measures the similarity using dot-products. The power normalisation unsparsifies the FV making it more suitable for linear SVMs.
The additional information provided by the means and the power normalisation explains the superior accuracy performance of FV.

\vspace{-0.5ex}
\subsection{Challenging Conditions}
\label{sec:chall_cond}
\vspace{-0.5ex}

In this section, we evaluate the performance on all datasets under consideration when the testing videos have translations and scale variations. 
We have selected the following approaches for this evaluation: $d_\text{spd}$, $K_\text{spd}^\text{poly}$, $K_\text{ls}^\text{rbf}$, and FV.  
We discard $d_\text{ls}$,  as its performance is too low and  presents similar behaviour to $d_\text{spd}$. 
We do not include experiments on KSR$_{\text{spd}}$ and KSR$_{\text{ls}}$, as we found that they show similar trends as  $K_\text{spd}^\text{poly}$ and $K_\text{ls}^\text{rbf}$, respectively.
Alike, GMM exhibits similar behaviour as FV.

\begin{figure}[!b]
  \vspace{-1ex}
  \centering
  
  \begin{minipage}{0.9\textwidth}
  
  \begin{minipage}{1\textwidth}
    \centering
    \begin{minipage}{0.32\textwidth}
      \centering
      \includegraphics[width=1\textwidth]{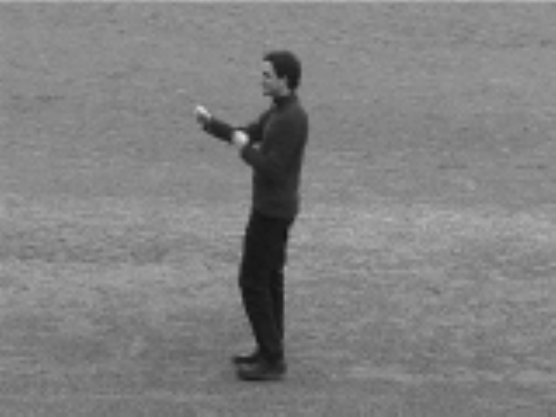}\\
      {\footnotesize original}
    \end{minipage}
    \begin{minipage}{0.32\textwidth}
      \centering
      \includegraphics[width=1\textwidth]{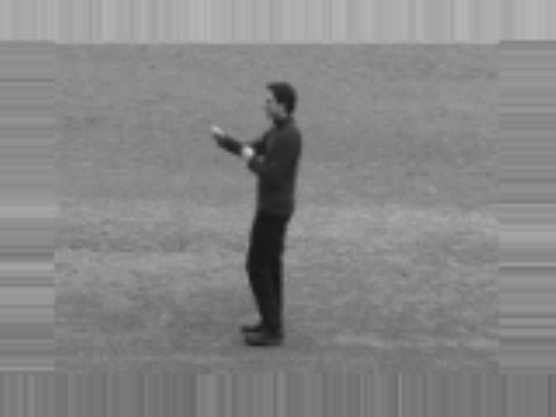}\\
      {\footnotesize scale: shrinkage}
    \end{minipage}
    \begin{minipage}{0.32\textwidth}
      \centering
      \includegraphics[width=1\textwidth]{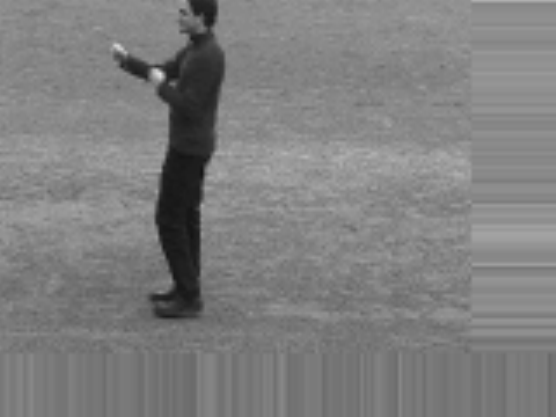}\\
      {\footnotesize translation: left and up}
    \end{minipage}
  \end{minipage}
  
  \end{minipage}
  
  \vspace{-1ex}
  \caption{Examples of challenging conditions.}
  \label{fig:chall_cond}
  
\end{figure}

For this set of experiments, the training is carried out using the original datasets.
For the analysis of translations, we have translated (shifted) each testing video vertically and horizontally. 
For the evaluation under scale variations, each testing video is shrunk or magnified.  
For both cases, we replace the missing pixels simply by copying the nearest rows or columns.
See Fig.~\ref{fig:chall_cond} for examples of videos under challenging conditions.

\begin{figure*}[tb!]
  \centering
  \begin{minipage}{1.0\textwidth}
    \centering
    \includegraphics[width=0.32\textwidth]{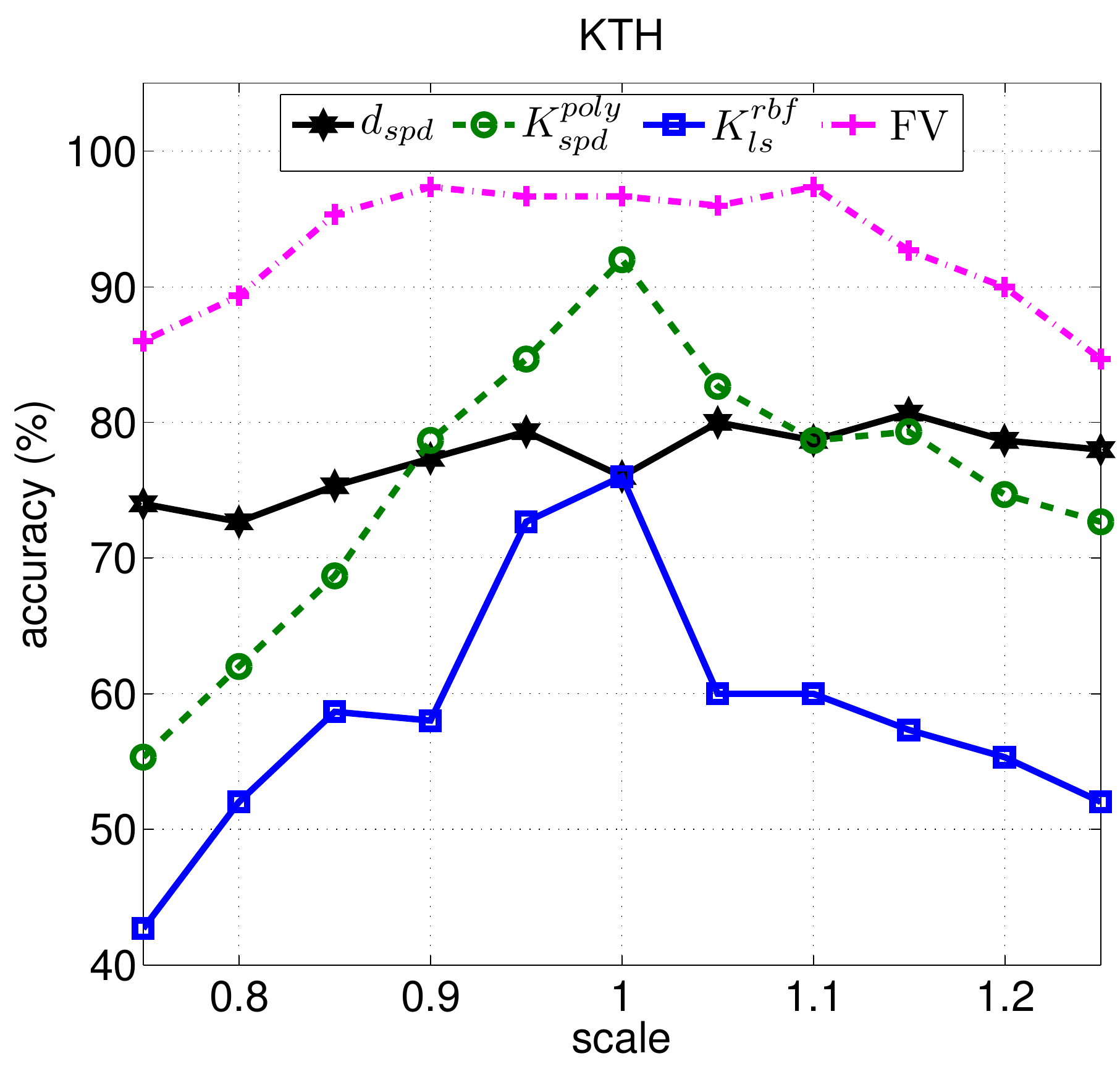}
    \includegraphics[width=0.32\textwidth]{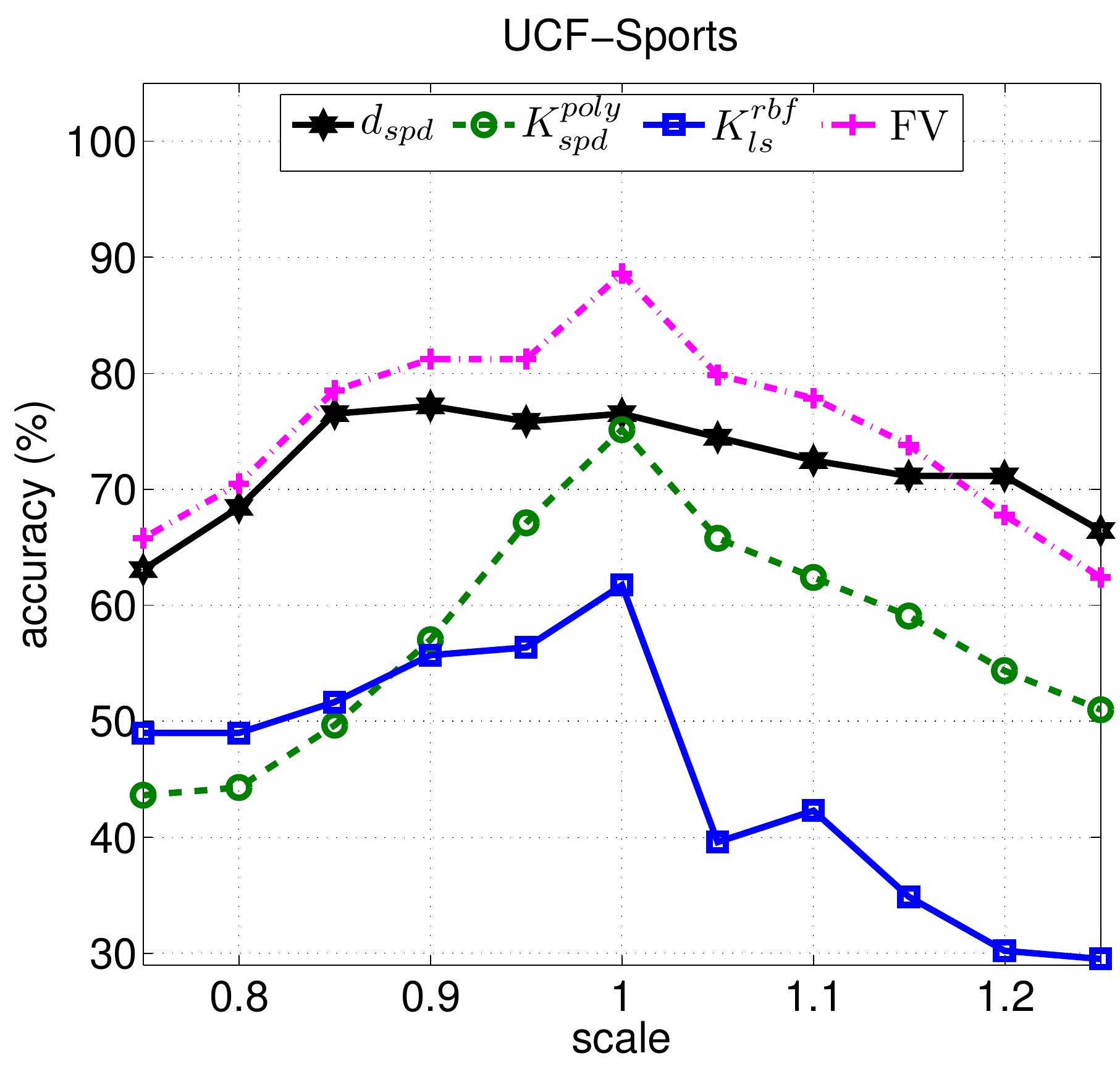}
    \includegraphics[width=0.32\textwidth]{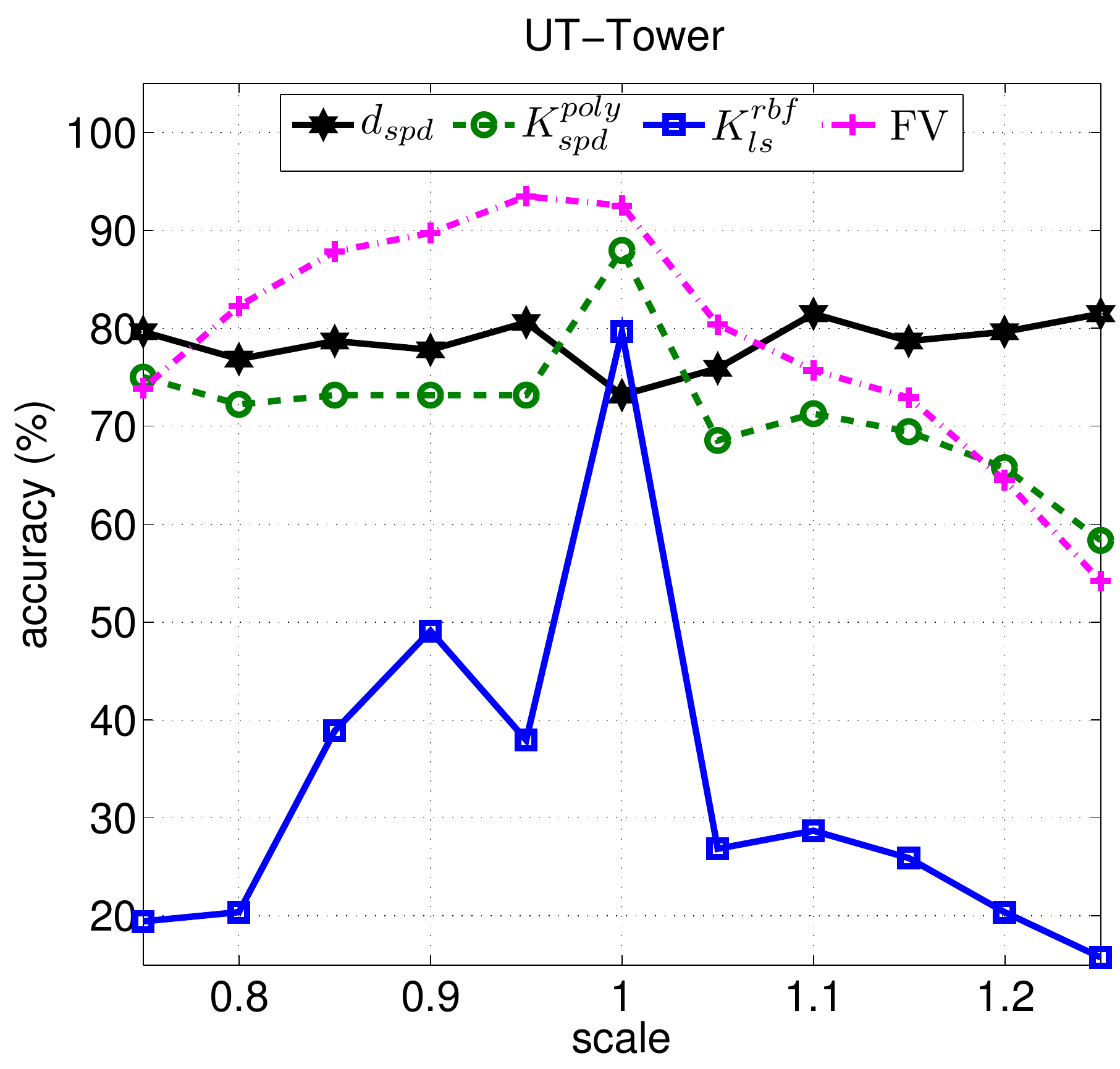}
  \end{minipage}
  
  \vspace{-2ex}
  \caption{Results for scale variation; scale \mbox{$>1$} means magnification, while \mbox{$<1$} means shrinkage.}
  \label{fig:scale_variation}
  
\end{figure*}

\begin{figure*}[tb!]
  \centering
  \begin{minipage}{1.0\textwidth}
    \includegraphics[width=0.32\textwidth]{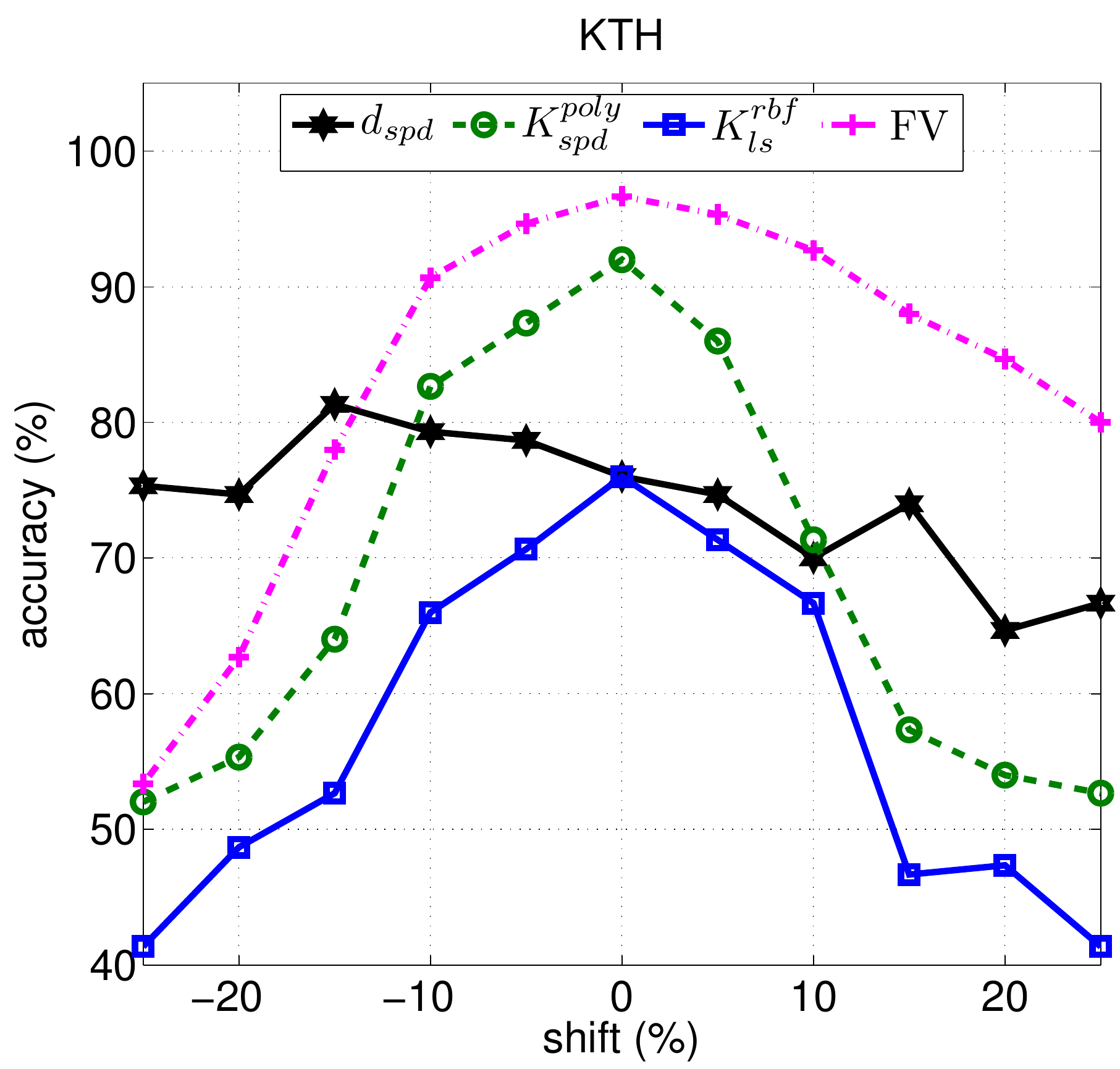}
    \includegraphics[width=0.32\textwidth]{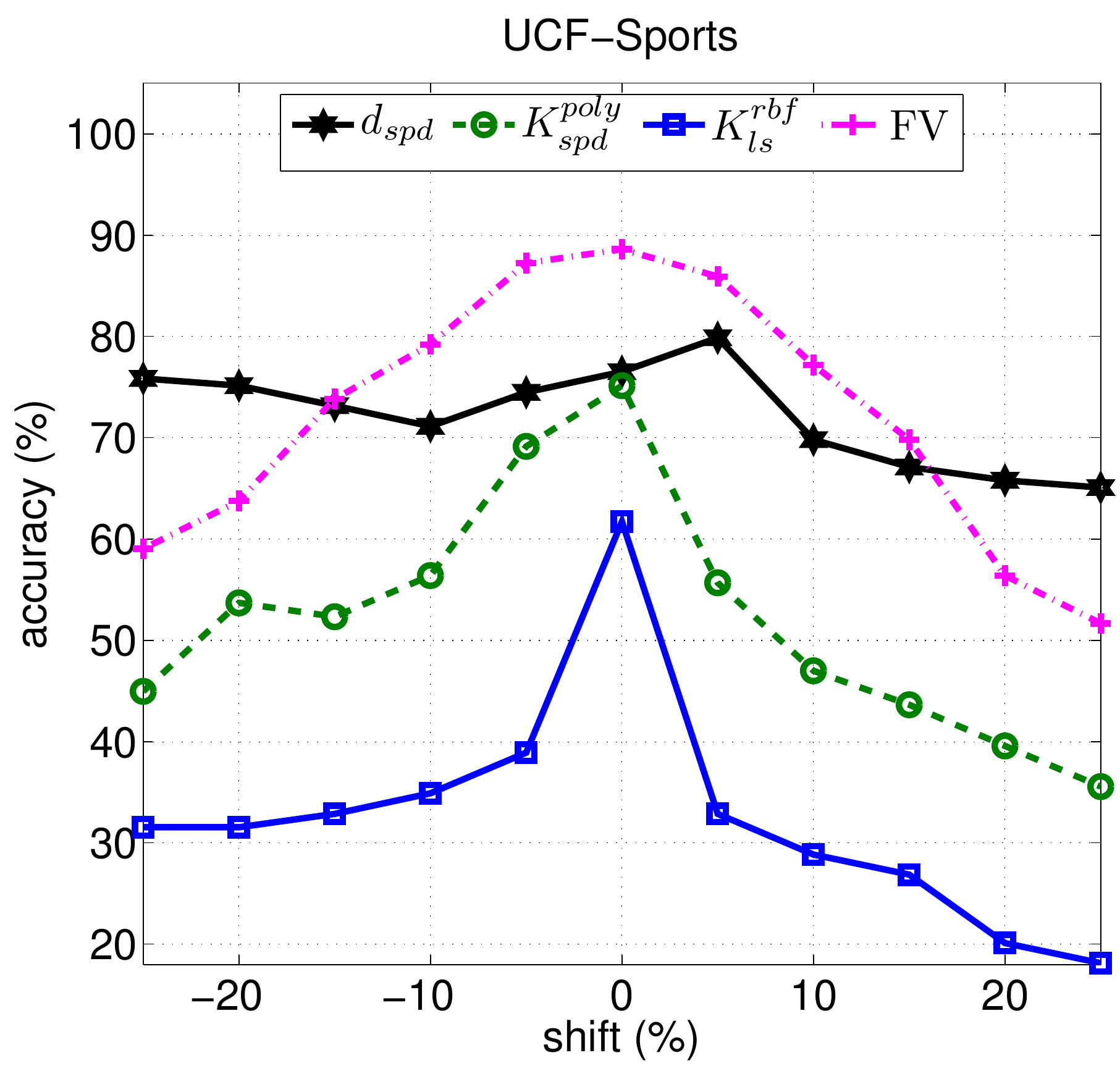}
    \includegraphics[width=0.32\textwidth]{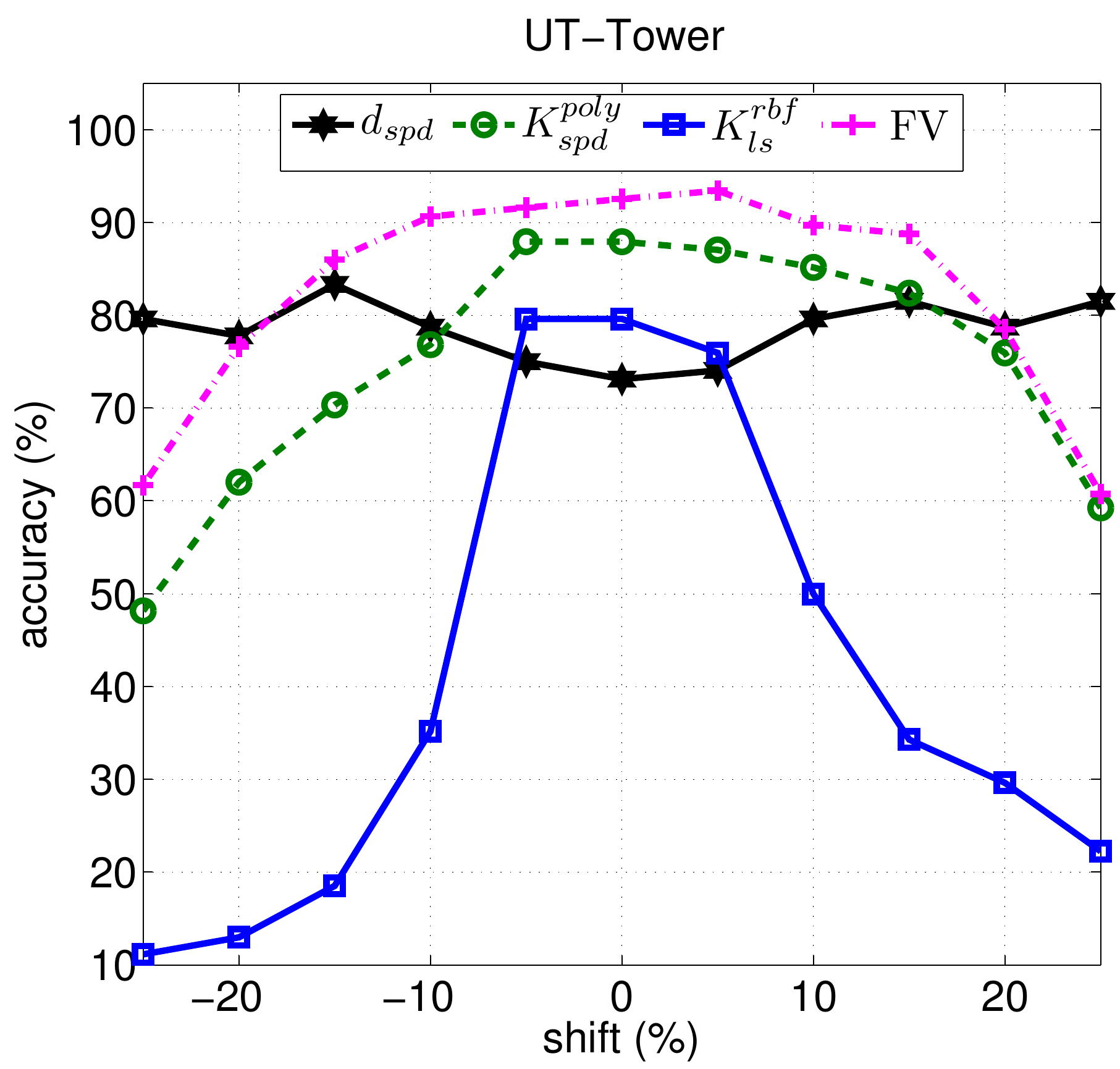}
  \end{minipage}
  
  \vspace{-2ex}
  \caption{Results for translation experiments. Each testing video is translated vertically and horizontally at the same time. A positive percentage indicates the video has been translated to the right and bottom while a negative percentage indicates the video has been translated to the left and up.}
  \label{fig:misalignments}
  
\end{figure*}

The results for scale variations and translations are shown in Figs.~\ref{fig:scale_variation} and ~\ref{fig:misalignments}, respectively. 
These results reveal that all the analysed approaches are susceptible to translation and scale variations. 
Both kernel based methods, $K_\text{spd}^\text{poly}$ and $K_\text{ls}^\text{rbf}$, exhibit  sharp performance degradation even when the scale is only magnified or compressed by a factor of $0.05\%$. Similarly, for both kernels the accuracy rapidly decreases with a small translation. 
The NN classification using the log-Euclidean distance ($d_\text{spd}$) is less sensitive to both variations. 
It can be explained by the fact that log-Euclidean metrics are by definition invariant by any translation and scaling in the domain of logarithms~\cite{Arsigny2006}. 
FV presents the best behaviour under  moderate variations in both scale and translation. 
We attribute this to the loss of explicit spatial relations between object parts.

\section{Main Findings}
\label{sec:conclusions}
\vspace{-1ex}

In this paper, we have presented an extensive empirical comparison  among existing techniques for the human action recognition problem. 
We have carried out our experiments using three popular datasets: KTH, UCF-Sports and UT-Tower. 
We have analysed Riemannian representations including  nearest-neighbour classification, kernel methods, and kernelised sparse representations. 
For Riemannian representation we used covariance matrices of features, which are symmetric positive definite (SPD),   as well as linear subspaces (LS). 
Moreover, we compared all the aforementioned Riemannian representations with GMM  and FV based representations, using the same extracted features.
We also evaluated the robustness of the most representative approaches to translation and scale variations. 

For manifold representations, all SPD matrices approaches surpass their LS counterpart, as a result of the use  of not only the dominant eigenvectors  but also  the eigenvalues.
The FV representation outperforms all the techniques under ideal and challenging conditions.
Under ideal conditions, FV achieves an overall accuracy of $92.6\%$, which is $7.6$ points higher than both GMM and the polynomial kernel using SPD matrices ($K_\text{spd}^{poly}$).
FV encodes more information than Riemmannian based methods, as it characterises the deviation from a probabilistic visual dictionary (a GMM) using means and covariance matrices.
Moreover, FV is less sensitive under moderate variations in both scale
and translation.

~

\begin{footnotesize}
\noindent
{\bf Acknowledgements.}
NICTA is funded by the Australian Government via the Department of Communications, and the Australian Research Council via the ICT Centre of Excellence~\mbox{program}.
\end{footnotesize}

\bibliographystyle{ieee}
\bibliography{references}

\end{document}